\documentclass{article}

\usepackage[preprint]{neurips_2024}

\usepackage[utf8]{inputenc} 
\usepackage[T1]{fontenc}    
\usepackage{hyperref}       
\usepackage{url}            
\usepackage{booktabs}       
\usepackage{amsfonts}       
\usepackage{nicefrac}       
\usepackage{microtype}      
\usepackage{xcolor}         
\usepackage{graphicx}

\title{Dynamic Mixture of Experts\linebreak Against Severe Distribution Shifts}

\author{%
  Donghu Kim \\
  Kim Jaechul Graduate School of AI \\
  KAIST \\
  \texttt{quagmire@kaist.ac.kr}
}

\begin{document}

\maketitle

\begin{abstract}
The challenge of building neural networks that can continuously learn and adapt to evolving data streams is central to the fields of continual learning (CL) and reinforcement learning (RL). This lifelong learning problem is often framed in terms of the plasticity-stability dilemma, focusing on issues like loss of plasticity and catastrophic forgetting. Unlike neural networks, biological brains maintain plasticity through capacity growth, inspiring researchers to explore similar approaches in artificial networks, such as adding capacity dynamically. Prior solutions often lack parameter efficiency or depend on explicit task indices, but Mixture-of-Experts (MoE) architectures offer a promising alternative by specializing experts for distinct distributions. This paper aims to evaluate a DynamicMoE approach for continual and reinforcement learning environments and benchmark its effectiveness against existing network expansion methods.
\end{abstract}

\section{Introduction}

Building a neural network that can indefinitely learn and adapt to an ever-changing stream of data is one of the core challenges in the machine learning (ML) community, especially continual learning (CL) and reinforcement learning (RL) fields. This \textit{lifelong learning} challenge can be decomposed into two smaller problems; loss of plasticity and catastrophic forgetting \cite{kumar2023continual}. This also manifests itself as the plasticity-stability dillema, which researchers refer to as catastrophic interference \cite{mccloskey1989catastrophic}. In this paper, we will primarily focus on addressing loss of plasticity while sidestepping the catastrophic forgetting problem.

Unlike neural networks, biological brains (especially those of humans) are capable of maintaining plasticity almost throughout their lifetime. One contributing factor is its \textit{growth} in capacity; human brains experience expansion in their cortical cortex to adapt to novel experiences \cite{Welker1990, hill2010similar}. Inspired by this phenomenon, many works proposed to constantly provide additional capacity to deal with loss of plasticity in neural networks. However, many prior works are proposed simply as a diagnostic tool (disregarding parameter-efficiency) \cite{nikishin2024deep}, or rely on explicit task index to deal with multiple distinct distributions \cite{malagonself, rusu2016progressive}.

Meanwhile, a huge potential for solving the problem can be found in the Mixture-of-Experts (MoE) literature~\cite{willi2024mixture, huang2024mentor}. The core idea that 'each expert specializes in handling distinct distributions' inspires the possibility that \textit{dynamically adding new experts into the MoE layer} can deal with distribution shift and provide additional plasticity to the network. This method, dubbed DynamicMoE, has the potential to constantly benefit from additional capacity while not relying on explicit task index. Thus, our objective in this paper would be the following:

\begin{itemize}
    \item Verify the effectiveness of dynamic MoE architecture in continual learning and reinforcement learning environments.
    \item Compare with prior network expansion works under a unified benchmark.
\end{itemize}

\section{Dynamic MoE}
Our method is conceptually simple; periodically add new experts to the MoE layers. We consider an architecture where any linear layer is included in one of the MoE layers. In order to keep the comparisons fair in terms of parameter count, we control the hidden dimension of each expert when increasing the number of experts. When naively adopted, this would limit the size of the feature space (Figure 1.a,b). Instead, we adopt the bottleneck expert architecture, where we control the inner hidden dimension of each expert but keep the input and output dimension consistent regardless of the number of experts (Figure 1.c,d). For simplicity, we do not consider function preservation when adding new experts \cite{gesmundo2023composable}. 

\begin{center}
    \begin{figure}[t]
    \begin{center}
    \includegraphics[width=0.85\linewidth]{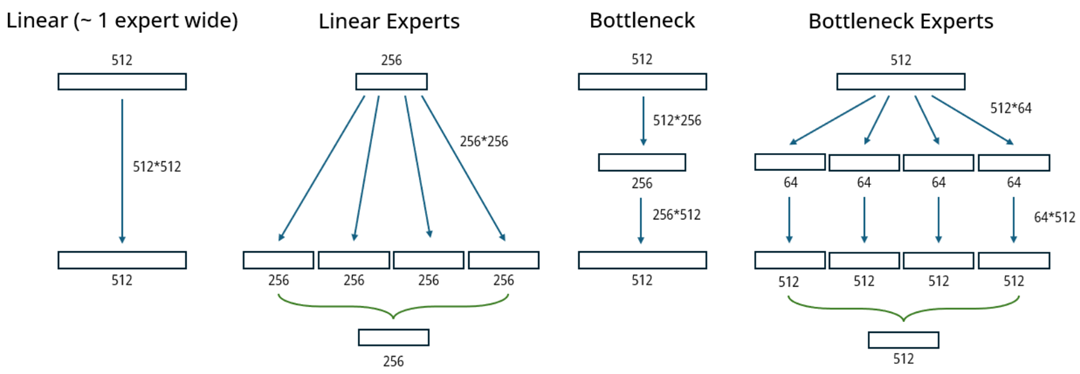}
    \end{center}
    \caption{An example diagram comparing (a) Linear layer, (b) Mixture of linear experts, (c) MLP with bottleneck layer, and (d) Mixture of bottlenecked experts. Ignoring router parameters and biases, all four networks have the same number of parameters.}
    \label{fig:overview}
    \end{figure}
\end{center}

\section{Experimental Results}

\subsection{Synthetic Continual Learning}
We first validate the effectiveness of Dynamic MoE by comparing with prior network expansion methods in a synthetic continual learning setup. We split the Tiny ImageNet dataset into 10 disjoint chunks, and train on each chunk for 50,000 steps while retaining previous chunks. This means that the whole dataset will be utilized in the final 50,000 steps.

We adopt a base model of 3-layer CNN with 3-layer MLP, and consider the following prior methods for expanding the MLP layers. We use each algorithm/method to expand the neural network at the start of new chunk (i.e., every 50,000 steps). We do not consider CNN layers.
\begin{itemize}
    \item \textbf{No Expansion}
    \item \textbf{Net2WiderNet} \cite{chen2015net2net} gradually increases the hidden dimension throughout the learning process. While the authors adopt a special initialization scheme, we used random initialization as it showed no significant performance loss.
    \item \textbf{Progressive Neural Network} \cite{rusu2016progressive} adds a new copy of the network that accepts the output of prior copies as input. Since we are considering a problem where no task index is defined, we only use final output of the lastest copy.
    \item \textbf{Plasticity Injection} \cite{nikishin2024deep} freezes current parameters and adds new randomly initialized parameters for training, while precisely conserving its function.
\end{itemize}
When using these methods, we control the hidden dimension of the whole network to match their parameters with the base model.

\clearpage

For Dynamic MoE, we consider three variants by their granularity:
\begin{itemize}
    \item \textbf{Granularity 1}: Starts with a single expert and adds one expert at every growth.
    \item \textbf{Granularity 2}: Starts with two experts and adds two experts at every growth. Each expert's inner hidden dimension is halved from \textbf{Granularity 1}.
    \item \textbf{Granularity 4}: Starts with four experts and adds four experts at every growth. Each expert's inner hidden dimension is halved from \textbf{Granularity 2}.
\end{itemize}
All algorithms are curated to end up with similar number of active parameters at its final training stage.

\begin{center}
    \begin{figure}[t]
    \begin{center}
    \includegraphics[width=0.85\linewidth]{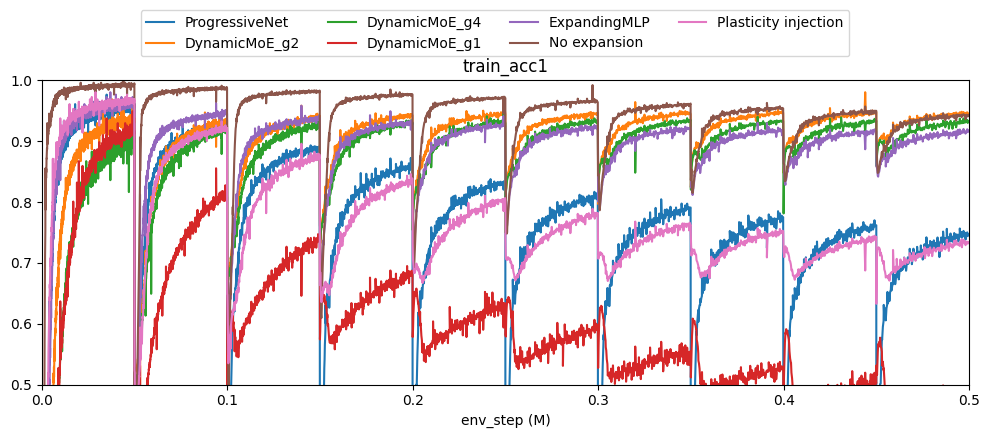}
    \end{center}
    \caption{Training accuracy of network expansion methods throughout the synthetic continual learning setup. \textbf{No expansion} starts with its maximal capacity, which explains its strong initial performance. However, it loses its trainability as shown by the degradation in performance. On the other hand, \textbf{DynaicMoE(g2, g4)} is able to maintain its initial performance until the final task. Other expansion methods did not exhibit this property. }
    \label{fig:synthetic}
    \end{figure}
\end{center}

Figure 2 shows the training accuracy throughout the learning process. We can see that although Dynamic MoE starts with lower performance, those with enough granularity (higher than 1) are able to almost perfectly maintain their initial performance until the last chunk. On the other hand, prior works and the base model (No Expansion) experience constant drop in their performance, eventually dropping lower than Dynamic MoE.

\subsection{Open World Environment}

We proceed to a more realistic setup, where we train an agent in an open world environment with severe distribution shifts via RL. We use the Pixel-based Craftax \cite{matthews2024craftax} environment, which presents more than 7 stages with unique visuals and tasks. We use PPO \cite{schulman2017proximal}, and leverage the state-of-the-art architecture SimBa \cite{lee2024simba} and convert each residual block into mixture-of-expert blocks. The residual blocks of SimBa are MLP structured, so we control the intermediate hidden dimension size similarly to Section 3.1. To verify the effectiveness of Dynamic MoE in environments with constant distribution shift, we consider the following baselines:

\begin{itemize}
    \item \textbf{No MoE}: No intervention is made to the network architecture, meaning the full capacity is utilized from the onset.
    \item \textbf{1 expert}: Trains only with single expert of the MoE layer.
    \item \textbf{K experts}: Trains with maximum experts from the start.
    \item \textbf{1-to-K grow}: Starts with a single expert, while periodically adding a new expert to each MoE layer K-1 times.
\end{itemize}

\begin{center}
    \begin{figure}[t]
    \begin{center}
    \includegraphics[width=0.85\linewidth]{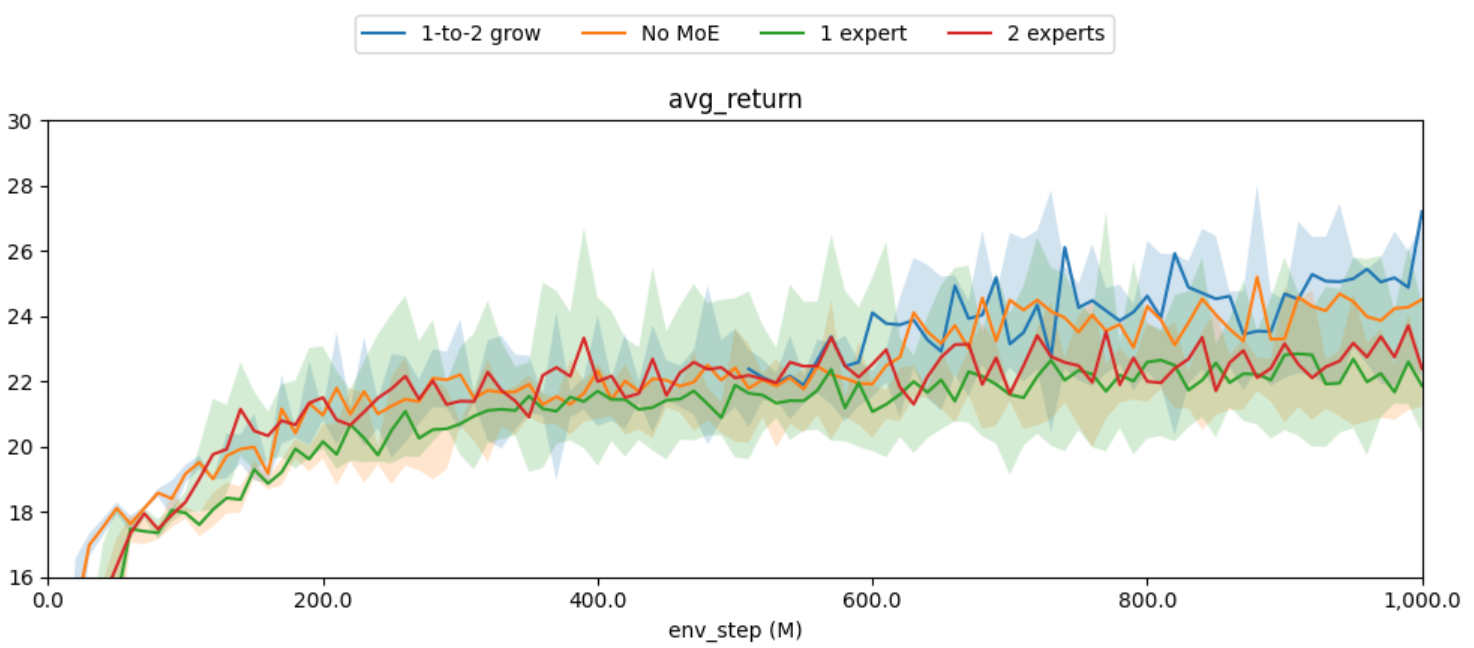}
    \end{center}
    \caption{1B Craftax results. We visualize the average performance and standard deviation over 3 random seeds. \textbf{1-to-2 grow} starts from the checkpoint created by \textbf{1 expert} to effectively see the divergence between adding new experts and not. }
    \label{fig:synthetic}
    \end{figure}
\end{center}

We consider the standard 1B environment steps, and use K=2. As a result, new experts are added once in 500M steps. Since there are total three MoE blocks (one in actor, two in critic), three experts are added to the architecture. Figure 3 draws the learning curve of each variants. Comparing 1 expert and 1-to-2 grow, we can see that there is a clear improvement in performance. Surprisingly, dynamically growing experts even surpasses the two baselines that use the full capacity from the start (No MoE, 2 experts). This could be explained that the capacity is wasted on overfitting early distributions, and that dynamically adding new experts is able to avoid this problem.

Further case study reveals an interesting property of Dynamic MoE. We visualize how the router weight changes within a episode in Figure 4, and find that distinct distributions (that arise in different stages of Craftax) are allocated to newly added experts. Such property was not found in other variants, meaning that dynamically adding new experts has a high potential to maintain plasticity even in continually shifting distributions.

\begin{center}
    \begin{figure}[t]
    \begin{center}
    \includegraphics[width=0.85\linewidth]{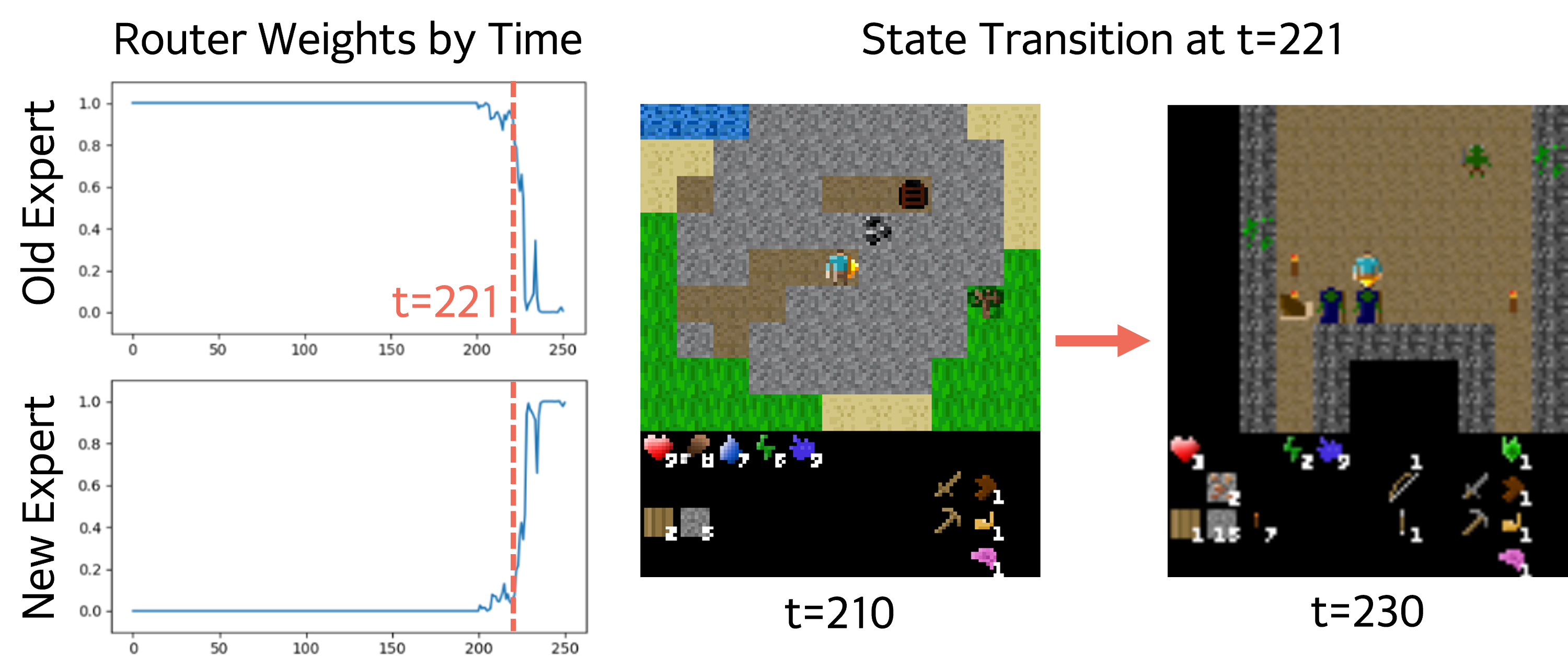}
    \end{center}
    \caption{(Left) Router weight visualization of the first MoE layer in the critic. As the agent enters second stage (dungeon) at $t=221$, the router weight promptly shifts from the old expert to the new one. (Right) Observation of Craftax as the stage transitions from first stage to the second stage. }
    \label{fig:synthetic}
    \end{figure}
\end{center}

\medskip

\acksection{We would like to express our gratitude to Clare Lyle, Hojoon Lee, and Johan Obando-Ceron for the helpful discussions and idea-sharing. }

\clearpage

{
\small

\bibliographystyle{plain}
\bibliography{reference} 

\end{document}